\documentclass[]{phai}

\usepackage[utf8]{inputenc}
\usepackage{amsmath}
\usepackage{amssymb}
\usepackage{amsfonts}
\usepackage{array}
\usepackage{colortbl}
\usepackage{nicefrac}
\usepackage{xspace}
\usepackage{url}

\newcommand{\modelname}{AnaDiffusion\xspace}

\title{AnaDiffusion: Anatomically Compositional Latent Diffusion for Controllable 3D Brain MRI Generation}

\author[1,2,\dagger]{Huiwen Han}
\author[2,3,\dagger]{Lulin Liu}
\author[4]{Bangya Liu}
\author[5]{Yuanhao Cai}
\author[2]{Nuo Chen}
\author[2]{Xiaoqing Wang}
\author[6]{Ziqian Xie}
\author[7]{Chenyu You}
\author[2]{Shuiwang Ji}
\author[6]{Degui Zhi}
\author[2]{Zhiwen Fan}

\affiliation[1]{Stanford University}
\affiliation[2]{Texas A\&M University}
\affiliation[3]{University of Minnesota}
\affiliation[4]{University of Wisconsin-Madison}
\affiliation[5]{Johns Hopkins University}
\affiliation[6]{Yale University}
\affiliation[7]{Stony Brook University}

\contribution[\dagger]{Equal contribution}

\abstract{
3D brain MRI generation has made significant advances in medical imaging, simulation, and controllable anatomical analysis. However, existing generative models typically synthesize 3D volumes monolithically, often overlooking regional anatomical structures and limiting local controllability. To address these limitations, we introduce \textbf{AnaDiffusion}, an anatomically compositional latent diffusion framework that factorizes the generation process into distinct, anatomically meaningful regions, followed by part-to-whole assembly and global refinement. Our approach first trains part diffusion models to capture local structural priors. We then inject an assembled anatomical composite of the parts into the whole-brain latent representation and continue denoising. This mechanism enables the model to resolve global context while preserving the injected anatomy. AnaDiffusion thus produces explicit part assets and a globally coherent volume, enabling controllable part editing without subject-specific dense segmentation maps at inference. On a subject-disjoint ADNI test split, AnaDiffusion attains the lowest mean Fr\'echet distance (FID) in every evaluated region. In localized editing, it achieves substantially higher target transfer (paired MS-SSIM) than a segmentation-conditioned baseline, with off-target preservation close to the baseline's.
}

\date{\today}
\metadata[Code]{\url{https://github.com/phai-lab/AnaDiffusion}}
\metadata[Project Page]{\url{https://anadiffusion.github.io/}}
\metadata[Affiliation Note]{Huiwen Han and Lulin Liu were visiting students at\\Texas A\&M University during this work.}

\begin{document}

\maketitle

\begin{figure*}[!htb]
  \centering
  \includegraphics[width=0.94\textwidth]{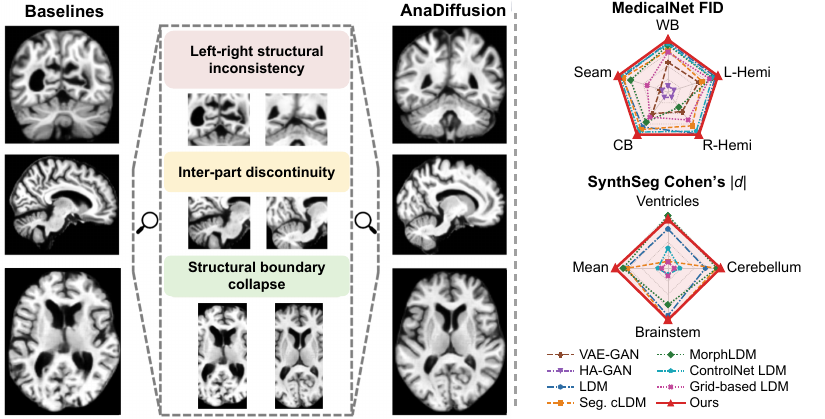}
  \caption{\textbf{AnaDiffusion for anatomically compositional 3D brain MRI generation.}
Unlike monolithic baselines (left) that may suffer from left-right inconsistency, inter-part discontinuity, and boundary collapse, AnaDiffusion generates part-specific anatomical assets and refines their assembled scaffold into a coherent whole-brain volume.
The radar charts (right) summarize Table 1: each axis is min--max normalized on a log scale and lower-is-better metrics (FID, |d|) are inverted, so points farther from the centre are better.}
  \label{fig:teaser}
\end{figure*}

\section{Introduction}
\label{sec:intro}
High-fidelity 3D brain MRI generation has become increasingly fundamental for medical imaging problems, where large-scale volumetric data are often costly to acquire, difficult to annotate, and limited by privacy and lack of paired anatomical segmentation. Brain MRI synthesis has benefited a wide range of downstream applications, including cross-modality translation \citep{kim2024adaptive}, super-resolution \citep{wang2023inversesr,khateri2025mri}, and disease progression simulation \citep{peng2023generating,dhinagar2024counterfactual,puglisi2024enhancing,fu2025synthesizing}. While recent diffusion models \citep{ho2020denoising,rombach2022high} have demonstrated remarkable success in synthesizing globally plausible volumetric data, generating anatomically accurate local structures remains a persistent challenge. Whole-brain samples that appear realistic at a macroscopic scale can exhibit localized anatomical failures--such as blurred tissue interfaces, topological inconsistencies in the cortex, and unnatural asymmetries--as shown in \Cref{fig:teaser}. These localized artifacts compromise the downstream utility of the generated data. In practice, they are often exposed by automated segmentation pipelines, which reveal incomplete or structurally inconsistent local anatomy in synthesized volumes \citep{wu2024evaluating,deo2025metrics,jafrasteh2025wasabi}.

These observations call for 3D MRI generators that preserve fine-grained anatomical structure along with global plausibility. Prior methods add explicit structural constraints. Geometry-driven approaches use deformation fields or surface meshes to encourage plausible shapes \citep{wang2025generating,wu2025igg,wilms2022invertible,bongratz2026cortex,rusak2022quantifiable}, and mask-based approaches use segmentation masks as conditioning signals or supervised objectives to enforce a prescribed spatial layout \citep{wan2026anatomically,konz2024anatomically,fernandez2024generating}. Mask conditioning gives spatial control but depends on dense input annotations, and the model learns to fill the given boundaries with intensities while the structural dependencies between adjacent regions remain implicit. Both mask-conditioned and unconditioned generators are also \emph{monolithic}: they encode and generate the entire volume through a single, entangled latent representation, although brain regions differ greatly in morphological complexity (e.g., the highly folded cerebral cortex and the densely structured cerebellum) \citep{fischl2002whole}. Monolithic models may therefore average out fine-scale, region-specific details, which motivates a framework that factorizes local structures and integrates them coherently.

Recognizing this gap, we introduce \textbf{AnaDiffusion}, an anatomy-informed compositional sampling and refinement framework for 3D brain MRI synthesis. Our contribution lies in how explicit regional generators are coupled to an ongoing whole-brain generative trajectory. To this end, we first train part-specific latent diffusion models, allowing the model to devote generative capacity to anatomically heterogeneous regions. For bilateral structures, we further employ a symmetry-aware hemisphere model with canonicalized weight sharing, which encourages anatomically consistent left-right generation. Building on these local anatomical priors, we then address the central challenge of part-to-whole integration. We construct an anatomical composite from the generated parts and use it to guide a global whole-brain diffusion process, which harmonizes regional details into a coherent volume. To preserve structural consistency, we train the whole-brain denoiser to resume reverse sampling from assembled composite latents reintroduced at intermediate noise levels. We summarize our contributions as follows:

\begin{itemize}
    \item We introduce \textbf{AnaDiffusion}, a compositional latent diffusion framework for controllable 3D brain MRI generation. AnaDiffusion decomposes synthesis into anatomical parts and a whole-brain latent integration process, modeling heterogeneous brain structures more effectively than monolithic whole-volume generation. 
\item \looseness=-1 We propose a \textbf{part-to-whole latent refinement} that integrates independently generated anatomical parts into a coherent whole-brain volume. This supports controllable part replacement while keeping non-edited regions stable and part-to-whole anatomy consistent.
    \item Motivated by the approximate structural bilateral symmetry of the human brain, we design a hemisphere generator shared across both hemispheres by employing left-right canonicalization and side-indicator conditioning. 
    
    \item We demonstrate that, relative to the compared baselines, AnaDiffusion improves the evaluated synthesis metrics on held-out subjects from a subject-disjoint ADNI test split. It achieves stronger regional distributional fidelity and segmentation-based anatomical alignment, while enabling localized part replacement with high target transfer and limited off-target drift, without requiring subject-specific dense segmentation maps at inference time.
\end{itemize}

\section{Related Work}
\label{sec:related_work}
\subsection{3D Generative Modeling for Brain MRI}
Early work generated brain MRIs slice-wise (2D) or by 2.5D aggregation; more recent work exploits improved compute and architectures for full 3D synthesis, which improves volumetric consistency. GAN-based models such as $\alpha$-GAN \citep{kwon2019generation,rosca2017variational}, CCE-GAN \citep{xing2021cycle}, and HA-GAN \citep{sun2022hierarchical} produce sharp 3D brain volumes, but their training is prone to instability and mode collapse, which matters when clinically relevant variation is subtle \citep{saad2024survey}. Diffusion models train more stably. 3D DDPM \citep{dorjsembe2022three} denoises directly in voxel space at a high computational cost, and latent diffusion models (LDMs) \citep{pinaya2022brain} lower this cost by running diffusion in the compact latent space of an autoencoder, which also makes conditioning straightforward \citep{peng2023generating,puglisi2024enhancing,herencia2025diffusion}. Anatomical plausibility and region-level control remain open problems for these models, and global perceptual metrics such as FID can miss local anatomical errors in generated brain volumes  \citep{wu2024evaluating,deo2025metrics,jafrasteh2025wasabi}. 
\subsection{Anatomy- and Morphology-Aware 3D Brain Generation}
To improve anatomical plausibility, recent 3D brain generators add explicit structural priors. Segmentation maps serve as conditioning signals or auxiliary supervision \citep{bhattacharya2025brainmrdiff, bongratz2026cortex, dorjsembe2024conditional}. For example, Med-DDPM \citep{dorjsembe2024conditional} concatenates segmentation masks with the image input to control normal and pathological brain generation, and AG-LDM \citep{wan2026anatomically} supervises both the autoencoder and the LDM with tissue- and boundary-aware losses. Morphology-aware methods add geometric constraints, for example template deformation in MorphLDM \citep{wang2025generating} and conditioned geodesic transformations for aging in IGG \citep{wu2025igg}. These priors regularize localization, tissue boundaries, or global shape, and each model still generates the brain as a single volume. AnaDiffusion decomposes the brain into anatomical parts with their own generators and assembles them in a whole-brain refinement step, which allows part-level editing without subject-specific dense segmentation maps at inference.

\section{Methods}
\label{sec:methods}

\begin{figure}[tb]
  \centering
  \includegraphics[width=\linewidth]{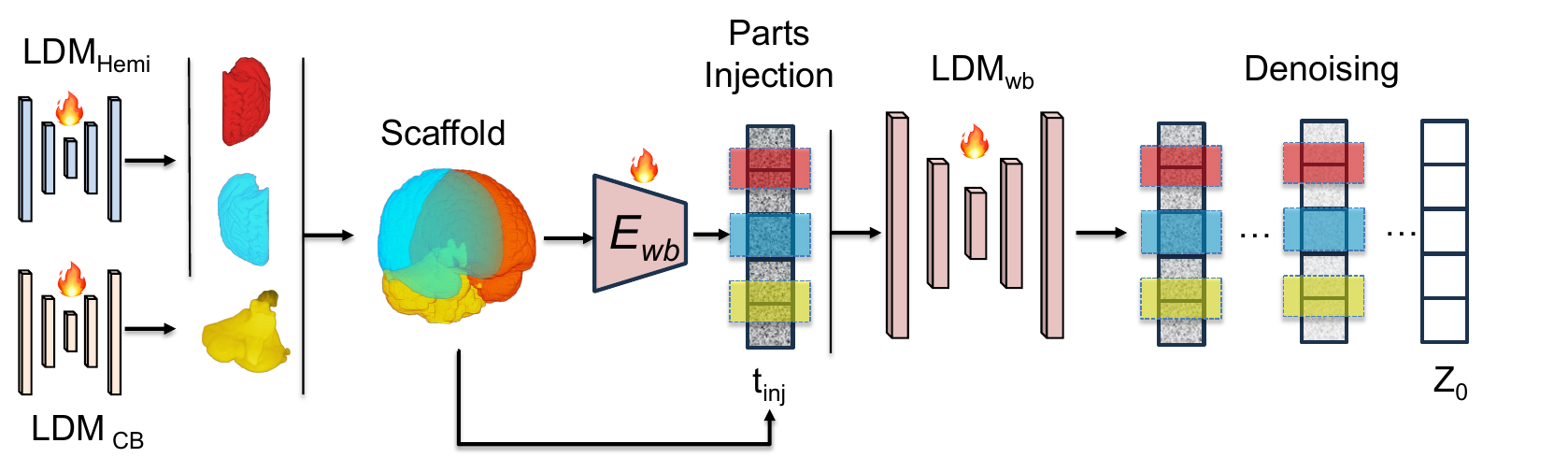}
  \caption{AnaDiffusion performs anatomically compositional 3D brain MRI generation by first synthesizing high-fidelity regional parts and assembling them into a whole-brain scaffold. 
The scaffold is re-encoded into the whole-brain latent space and injected during denoising, enabling local anatomical detail and globally coherent whole-brain synthesis.
  }
  \label{fig:example}
\end{figure}

\paragraph{\textbf{Overview.}}
We introduce \modelname, a part-to-whole latent diffusion framework for compositional 3D brain MRI generation. AnaDiffusion trains an LDM for each anatomical part and one for the whole brain (\Cref{fig:example}), then optimizes only the whole-brain denoiser with an assemble-then-refine strategy. Frozen part models generate anatomical parts that are assembled in image space, re-encoded, and injected as a composite latent at $t_{\mathrm{inj}}$. Training optimizes the whole-brain denoiser with standard \(\epsilon\)-prediction on scaffold-injected latents, together with a seam-band loss that anchors part boundaries to the real scan. At inference, the model produces explicit part assets and a whole-brain volume, enabling part replacement without subject-specific dense segmentation maps. 

\subsection{Anatomical Part Factorization}
\label{sec:part_factorization}
Let \(x^{\mathrm{wb}}\in\mathbb{R}^{H\times W\times D}\) denote a whole-brain T1 MRI registered to a common template space. We define \(P=3\) anatomical parts: left hemisphere (\(\text{Hemi}_L\)), right hemisphere (\(\text{Hemi}_R\)), and cerebellar-brainstem complex (\(CB\)). This coarse decomposition follows widely used neuroanatomical organization and is broadly consistent with anatomical and developmental perspectives~\citep{liang2007automatic,gui2012morphology}. Let \(M^{(p)}\in\{0,1\}^{H\times W\times D}\) be the binary mask for part \(p\), and let \(x^{(p)}=\mathcal{C}_p(x^{\mathrm{wb}})=M^{(p)}\odot x^{\mathrm{wb}}\) denote the corresponding part volume. During training-data preparation, part volumes are extracted from each whole-brain volume by applying part-specific segmentation label maps generated with SynthSeg~\citep{billot2023synthseg}, and then cropped to fixed, part-aligned bounding boxes. These subject-derived labels are used only offline to construct the regional training volumes and are not provided to the diffusion models as subject-specific conditioning inputs. Extra-ventricular cerebrospinal fluid (CSF) is not assigned to any part because it is a fluid compartment whose boundaries are unstable and sensitive to partial-volume effects. We define a deterministic assembly operator \(x^{\mathrm{scaf}}=\mathcal{A}(\{x^{(p)}\}_{p=1}^{P})\), which places part assets into their template-aligned locations while omitting CSF, background, and uncertain boundary regions. Details on the anatomical definitions, template masks, crop geometry, and assembly procedure are provided in Appendix~\ref{sec:supp_parts}.

\subsection{Local Anatomical Priors}
\label{sec:local_priors}

For each anatomical support, we train an autoencoder and latent diffusion prior. Let \(E^{\mathrm{wb}},D^{\mathrm{wb}}\) denote the whole-brain encoder and decoder, and \(E^{(p)},D^{(p)}\) denote the corresponding part autoencoder. All autoencoders are trained before diffusion training and then frozen. Given a latent \(z_0=E(x)\), each diffusion denoiser is trained with the standard noise-prediction objective
\begin{equation}
    \mathcal{L}_{\epsilon}
    =
    \mathbb{E}_{z_0,t,\epsilon}
    \left[
        \left\|
        \epsilon - \epsilon_{\theta}(z_t,t)
        \right\|_2^2
    \right],
    \qquad
    z_t = \alpha_t z_0 + \sigma_t \epsilon,
    \quad
    \epsilon\sim\mathcal{N}(0,I).
    \label{eq:eps_loss}
\end{equation}

\paragraph{\textbf{Hemisphere canonicalization.}}
The cerebral hemispheres have approximate bilateral correspondence but also contain side-specific variation. We therefore use canonicalization for statistical sharing, not exact symmetry enforcement. Let \(x^{L}\) and \(x^{R}\) denote left- and right-hemisphere crops. We map right-hemisphere crops into a left-oriented canonical frame using a flip operator \(\mathcal{F}\), i.e., \(\bar{x}^{R}=\mathcal{F}(x^{R})\). A single hemisphere LDM is trained on \(\{x^{L}\}\cup\{\bar{x}^{R}\}\) and conditioned on a side indicator \(y\in\{0,1\}\), where \(y=0\) denotes left and \(y=1\) denotes right. At inference, the model samples in the canonical frame; right-hemisphere samples are flipped back to native orientation using \(\mathcal{F}\).

\subsection{Part-to-Whole Latent Scaffold}
\label{sec:latent_scaffold}
Independently generated parts live in part-specific latent spaces and need not be directly compatible with the whole-brain latent space. We therefore compose in image space and re-encode with the whole-brain encoder. Given generated part assets \(\{\tilde{x}^{(p)}\}_{p=1}^{P}\), we assemble \(\tilde{x}^{\mathrm{scaf}}=\mathcal{A}(\{\tilde{x}^{(p)}\}_{p=1}^{P})\), then map it into the whole-brain latent space as \(z^{\mathrm{scaf}}_0 = E^{\mathrm{wb}}(\tilde{x}^{\mathrm{scaf}})\). This avoids direct pasting between incompatible part and whole-brain latent spaces. Each part is pasted into whole-brain coordinates with a fixed map, built from its MNI152 template mask and feathered at the boundary (Appendix~\ref{sec:supp_assembly}). Let \(m\) be the downsampled latent-space union of these part-masks in whole-brain coordinates. For a whole-brain latent trajectory \(z^{\mathrm{wb}}_t\), we noise the scaffold latent with the trajectory's own noise at the injection step and replace the content inside the part-masks with the scaffold,
\begin{equation}
z^{\mathrm{inj}}_{t_{\mathrm{inj}}}=z^{\mathrm{wb}}_{t_{\mathrm{inj}}}+\alpha_{t_{\mathrm{inj}}}\,m\odot\big(z^{\mathrm{scaf}}_0-\hat z_0\big),
\qquad
\hat z_0=\frac{z^{\mathrm{wb}}_{t_{\mathrm{inj}}}-\sigma_{t_{\mathrm{inj}}}\,\epsilon_\theta\big(z^{\mathrm{wb}}_{t_{\mathrm{inj}}},t_{\mathrm{inj}}\big)}{\alpha_{t_{\mathrm{inj}}}}.
\label{eq:latent_injection}
\end{equation}
The whole-brain denoiser then continues reverse sampling from \(z^{\mathrm{inj}}_{t_{\mathrm{inj}}}\) to obtain the final latent \(\hat{z}^{\mathrm{wb}}_0\), which is decoded as \(\hat{x}^{\mathrm{wb}}=D^{\mathrm{wb}}(\hat{z}^{\mathrm{wb}}_0)\).
\subsection{Assemble-then-Refine Compositional Training}
\label{sec:assemble_refine}

We train the whole-brain denoiser to refine scaffold-injected latents while keeping all part pipelines and the whole-brain autoencoder frozen. For a training scan \(x^{\mathrm{wb}}\), let \(z^{\mathrm{tgt}}_0=E^{\mathrm{wb}}(x^{\mathrm{wb}})\) be the target whole-brain latent. Frozen part generators produce part assets, which are assembled into \(\tilde{x}^{\mathrm{scaf}}\) and re-encoded as \(z^{\mathrm{scaf}}_0\). We sample \(t=t_{\mathrm{inj}}\sim\mathcal{U}[100,300]\), noise the target as \(z_t=\alpha_t z^{\mathrm{tgt}}_0+\sigma_t\epsilon\), inject the scaffold with Eq.~\eqref{eq:latent_injection} using the stop-gradient estimate \(\hat z_0\), and minimize
\begin{equation}
\mathcal{L}_{\mathrm{AR}}=\mathcal{L}_{\epsilon}+\lambda\,\mathbb{E}\Big[w_t\big\|M_{\mathrm{seam}}\odot\big(\epsilon^{*}-\epsilon^{\mathrm{wb}}_{\theta}(z^{\mathrm{inj}}_t,t)\big)\big\|_2^2\Big],
\qquad
\epsilon^{*}=\frac{z^{\mathrm{inj}}_t-\alpha_t z^{\mathrm{tgt}}_0}{\sigma_t},
\label{eq:ar_loss}
\end{equation}
where \(\mathcal{L}_{\epsilon}\) is the noise-prediction loss of Eq.~\eqref{eq:eps_loss}, \(M_{\mathrm{seam}}\) is a band around the part boundaries in latent space, \(w_t=\min(1,\sigma_t^2/\alpha_t^2)\), and \(\lambda=0.5\). The seam term asks the denoiser to recover the real scan across part boundaries, where the scaffold and the surrounding brain disagree.
This objective trains the whole-brain denoiser to continue reverse sampling from a latent state that already contains an assembled anatomical scaffold. The injected scaffold provides explicit regional structure, while the whole-brain denoiser synthesizes omitted regions such as CSF, background, and uncertain boundaries and harmonizes the final global anatomy.
\subsection{Part Editing Application}
\label{sec:inference_editing}

\looseness=-1 At inference, \modelname supports two modes. In \emph{unconditional compositional generation}, AnaDiffusion first denoises a whole-brain latent from Gaussian noise to the injection point \(t_{\mathrm{inj}}\). The part branches start from the whole-brain sample. We decode the partially denoised whole-brain estimate and use fixed MNI152-derived template masks to localize and crop anatomical regions from it, so each part refinement starts from a globally contextualized coarse structure. Each crop is refined into an explicit part asset by latent-space SDEdit. It is encoded with its part autoencoder, forward-noised to the injection timestep, and denoised by the frozen part LDM. The assets are pasted with the part masks of Section 3.3 and assembled into \(\tilde{x}^{\mathrm{scaf}}\) and re-encoded with \(E^{\mathrm{wb}}\), re-noised to \(t_{\mathrm{inj}}\) via Eq.~\eqref{eq:latent_injection}, and further denoised to obtain the final coherent volume. No subject-specific dense segmentation map is used at inference.

Part editing is intended for controllable localized synthesis, such as data augmentation for under-represented regional phenotypes (e.g., unilateral hippocampal atrophy) and counterfactual analyses; clinically specific edits would require regional generators trained with the corresponding condition. In \emph{part editing}, we first generate or select an initial whole-brain sample and its associated part assets. A selected asset, such as the right hemisphere or cerebellar-brainstem complex, may be replaced with an alternative generated part. The modified asset set is then reassembled, re-encoded, injected, and refined by the whole-brain denoiser. This yields a localized edit while preserving the non-target anatomy and repairing seams around the replacement.

During DDIM inference, \(r_{\mathrm{inj}}\) denotes the number of denoising steps remaining after scaffold injection, which is distinct from the training timestep \(t_{\mathrm{inj}}\). Smaller \(r_{\mathrm{inj}}\) preserves the inserted part more strictly, and larger \(r_{\mathrm{inj}}\) allows more harmonization (\Cref{sec:ablation}).

\section{Data Preprocessing and Anatomical Factorization}
\label{sec:data_prep_and_part_def}

\paragraph{\textbf{Dataset and preprocessing.}}
\begingroup
\renewcommand{\thefootnote}{\fnsymbol{footnote}}
\looseness=-1 Data used in the preparation of this article were obtained from the Alzheimer's Disease Neuroimaging Initiative (ADNI) database (\url{adni.loni.usc.edu}), and Appendix~\ref{sec:supp_adni} gives the full ADNI statement. We evaluate on 1,735 T1-weighted brain MRIs from 407 subjects. Subjects are 55-93 years old (mean 76.7), with 47.4\% female
subjects and diagnostic groups spanning cognitively normal, mild cognitive
impairment, and Alzheimer's disease. Each scan is preprocessed with N4 bias-field
correction~\citep{tustison2010n4itk}, skull stripping~\citep{hoopes2022synthstrip},
affine registration to MNI152 space~\citep{avants2008symmetric,iglesias2023ready}\footnote[1]{We used the Montreal Neurological Institute MNI152 template for image processing purposes; it is available for download at \url{http://www.bic.mni.mcgill.ca/ServicesAtlases/ICBM152NLin2009}.},
resampling to \(1.5\,\mathrm{mm}\) isotropic resolution, and intensity
normalization~\citep{shinohara2014statistical}.
To avoid leakage across longitudinal
acquisitions, we split the data at the subject level into training, validation,
and test sets containing 305, 41, and 61 subjects and 1,288, 156, and 291 scans,
respectively. AnaDiffusion was trained on the ADNI training split and evaluated
on the subject-disjoint ADNI test split.
\par
\endgroup

\paragraph{\textbf{Anatomical factorization.}}
We use a coarse, task-driven anatomical factorization into three components:
the left hemisphere (\(\text{Hemi}_\text{L}\)), the right hemisphere (\(\text{Hemi}_\text{R}\)), and the
cerebellar-brainstem complex (\(\text{CB}\)). This decomposition follows widely used neuroanatomical organization in neuroimaging and neurobiology and is broadly consistent with anatomical and developmental perspectives \citep{liang2007automatic, gui2012morphology}. The three parts are interpretable compartments with distinct morphology, the two hemispheres share homologous structure that canonicalization can exploit, and each part is a meaningful unit for localized editing. A recent sparse mixture-of-experts foundation model for multimodal brain MRI also learned experts that specialize to spatially distinct anatomical structures \citep{huang2026neurojepa}, which gives independent support for allocating model capacity by anatomy. SynthSeg labels are used only offline to build the regional training volumes (\Cref{sec:part_factorization}). Because all volumes are aligned to MNI152 space, fixed template masks give the crop locations and the paste weights at inference, so assembly needs neither subject-specific dense segmentation maps nor additional registration.

\section{Experiments}

\subsection{Experimental Setting}

\paragraph{\textbf{Baselines.}} We compare AnaDiffusion against seven representative baselines spanning adversarial, diffusion-based, and morphology-based generation: \textbf{(i) VAE-GAN} \citep{larsen2016autoencoding}, which combines variational autoencoding with adversarial training; \textbf{(ii) HA-GAN} \citep{sun2022hierarchical}, a hierarchical adversarial model for high-resolution 3D medical image synthesis; \textbf{(iii) a standard whole-brain LDM} \citep{pinaya2022brain,pinaya2023generative}, trained unconditionally; \textbf{(iv) a 3D ControlNet-style LDM} with a matched whole-brain backbone; \textbf{(v) a grid-based LDM} that partitions the volume into a $3\times3\times3$ grid and applies positional encoding to each of the resulting 27 patches, designed to separate the effect of anatomical semantics from generic spatial specialization; \textbf{(vi) a segmentation-mask cLDM}, representing prior mask-guided medical diffusion models (e.g., Med-DDPM \citep{dorjsembe2024conditional}), which conditions generation on anatomical part masks through multi-channel concatenation; and \textbf{(vii) MorphLDM} \citep{wang2025generating}, a template-warping approach that generates subject-specific anatomy by learning deformation fields relative to a canonical template. All baselines use identical subject-level splits and preprocessing, with matched whole-brain backbone configurations where applicable. MorphLDM is the sole preprocessing exception because it requires intensity normalization to $[0,1]$ rather than $[-1,1]$.
 
\paragraph{\textbf{Implementation.}} \looseness=-1 We implement AnaDiffusion in PyTorch using MONAI Generative Models AutoencoderKL
and diffusion UNet architectures~\citep{pinaya2023generative,cardoso2022monai}.
Generation resolutions are \(128^3\) for whole brain, \(64\times128\times128\)
for each hemisphere, and \(128\times96\times64\) for the cerebellar--brainstem
complex. Autoencoders are trained with AdamW for up to 100 epochs and then frozen
for diffusion training. We use a 1,000-step DDPM process with a scaled-linear
\(\beta\) schedule \((\beta_{\mathrm{start}}=0.0015,\beta_{\mathrm{end}}=0.0195)\)
and 50-step DDIM sampling at inference. During training, scaffold injection uses
\(t_{\mathrm{inj}}\sim\mathcal{U}[100,300]\) on the DDPM schedule. At inference,
we inject with \(r_{\mathrm{inj}}=10\) DDIM steps remaining, which corresponds to
approximately \(t_{\mathrm{inj}}\approx180\) under the 50-step DDIM subsampling (Appendix~\ref{sec:supp_training}).

\paragraph{\textbf{Metrics.}} We evaluate generation quality with the Fr\'echet distance (FID) in the feature space of a 3D ResNet-10 backbone from MedicalNet \citep{chen2019med3d}, pre-trained on large-scale medical imaging datasets. FIDs are computed for the whole brain and for template-defined subregions: left hemisphere, right hemisphere,
and cerebellar-brainstem complex. Subregion masks are derived from an MNI-space, SynthSeg template and dilated by 2 voxels (\(\approx3\) mm at 1.5 mm
isotropic spacing) to encompass residual anatomical size variation after registration.
To quantify interface coherence, we compute the same FID within a fixed seam band
around part boundaries. Because FID can miss local anatomical errors, we also assess anatomical plausibility with
SynthSeg-derived morphometry by comparing real and generated per-structure volume
distributions using Cohen's \(|d|\)~\citep{wu2024evaluating}. Each method generates 291 volumes, matching the 291 real test volumes. We report bootstrap means and 95\% percentile confidence intervals over 5,000 replicates, each resampling test subjects with replacement (keeping all their longitudinal scans) and, independently, each method's generated volumes.

\subsection{Evaluation of Generation Quality}

\begin{figure}[tb]
  \centering
  \includegraphics[width=\linewidth]{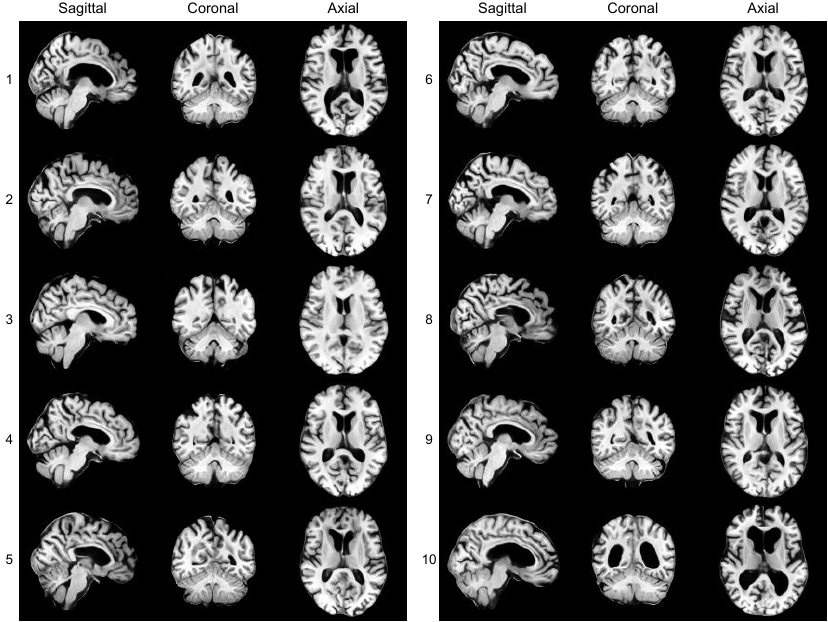}
  \caption{\textbf{Additional AnaDiffusion samples.}
  Ten representative generated samples, each shown as sagittal, coronal and axial slices (slice indices 59, 40 and 65).}
  \label{fig:additional_samples}
\end{figure}

\looseness=-1 Table~\ref{tab:main_metrics} shows that AnaDiffusion obtains the lowest mean MedicalNet FID across the whole brain, both hemispheres, the cerebellar--brainstem complex, and seam regions, suggesting improved distributional alignment at both global and part-specific scales. The seam-region gain is consistent with the intended role of whole-brain refinement in harmonizing independently generated parts. For SynthSeg-derived Cohen's \(|d|\), AnaDiffusion achieves the lowest cerebellar mismatch, \(0.169\,[0.007,0.482]\), and the second-lowest ventricular and brainstem mismatch, \(0.144\,[0.005,0.416]\) and \(0.185\,[0.008,0.483]\), respectively. These structures cover complementary aspects of anatomy. Ventricles are sensitive to neurodegenerative morphology and lie between parts, while cerebellum and brainstem contain compact structures with fine tissue boundaries and correspond to the dedicated \(CB\) part and its boundary with the cerebrum. SynthSeg label maps of AnaDiffusion samples also agree better with real test volumes than those of the LDM baseline in 16 of 17 regions for both Dice and HD95 (Appendix~\ref{sec:supp_structural_agreement}, \Cref{tab:structural_agreement}), and \Cref{fig:additional_samples} shows additional samples.

\begin{figure}[tb]
  \centering
  \includegraphics[width=\linewidth]{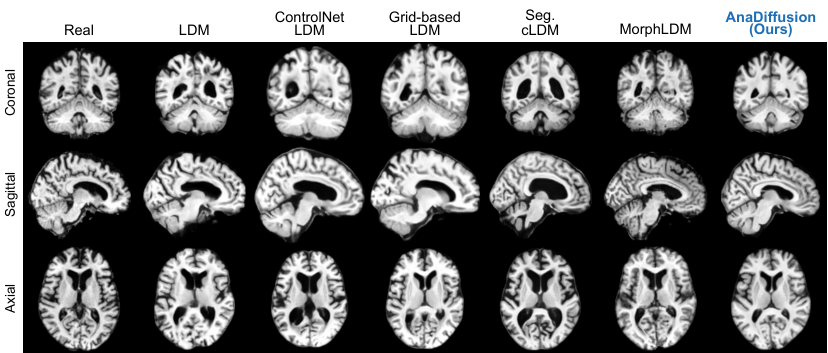}
  \caption{\textbf{Qualitative comparison of 3D brain MRI generation.} Columns from left to right show Real, LDM, ControlNet LDM, Grid-based LDM, Seg. cLDM, MorphLDM, and AnaDiffusion (Ours). Monolithic baselines show localized artifacts and lose fine-scale structure, whereas part-to-whole refinement preserves local anatomy and boundary coherence.}
  \label{fig:ablation}
\end{figure} 

\begin{table}[tb]
  \centering
  \small
  \renewcommand{\arraystretch}{1.05}
  \setlength{\tabcolsep}{1.5pt}
  \caption[Generation and focused morphometric metrics on ADNI.]{
  \textbf{Generation and focused morphometric metrics on \textit{ADNI}.}
  MedicalNet~\citep{chen2019med3d} FID is reported as FID $\times 10^4$. SynthSeg \citep{billot2023synthseg} metrics report absolute Cohen's
  $|d|$ for structures most directly tied to part-to-whole integration.
  WB: whole brain. Hemi: hemisphere. CB: cerebellar-brainstem complex.
  Seam FID is computed within a fixed seam band. 
  Values are reported as mean [95\% percentile bootstrap confidence interval; Section 5.1], and rankings are based on the means.
  Best results are in \textbf{bold}; second-best results are \underline{underlined}. Lower is better.}
  \label{tab:main_metrics}

  \newcommand{\metricci}[3]{%
    \shortstack{#1\\[-0.15em]{\scriptsize [#2, #3]}}%
  }

  \resizebox{\textwidth}{!}{%
  \begin{tabular}{@{}lccccc@{\hspace{4pt}}ccc@{}}
    \toprule
    \multirow[c]{2}{*}[-0.35ex]{Method}
    & \multicolumn{5}{c}{MedicalNet FID}
    & \multicolumn{3}{c}{SynthSeg Cohen's $|d|$} \\
    \cmidrule(lr){2-6} \cmidrule(l){7-9}
    & WB & Left Hemi & Right Hemi & CB & Seam
    & Ventricles & Cerebellum & Brainstem \\
    \midrule
    VAE-GAN~\citep{larsen2016autoencoding}
    & \metricci{103.54}{75.1}{140.0}
    & \metricci{17.21}{13.4}{23.1}
    & \metricci{15.35}{11.1}{21.4}
    & \metricci{3.73}{2.8}{5.5}
    & \metricci{16.79}{12.8}{22.3}
    & -- & -- & -- \\

    HA-GAN~\citep{sun2022hierarchical}
    & \metricci{310.33}{262.9}{360.3}
    & \metricci{63.67}{55.7}{72.0}
    & \metricci{25.75}{22.0}{30.0}
    & \metricci{9.87}{8.3}{11.6}
    & \metricci{20.09}{17.1}{23.3}
    & -- & -- & -- \\

    LDM~\citep{pinaya2022brain}
    & \metricci{\underline{40.92}}{26.71}{58.86}
    & \metricci{\underline{7.10}}{4.26}{10.85}
    & \metricci{\underline{6.87}}{4.51}{9.57}
    & \metricci{1.68}{1.17}{2.29}
    & \metricci{0.44}{0.28}{0.63}
    & \metricci{0.184}{0.007}{0.523}
    & \metricci{0.370}{0.036}{0.747}
    & \metricci{0.222}{0.011}{0.540} \\

    Seg. cLDM~\citep{dorjsembe2024conditional}
    & \metricci{59.47}{32.72}{90.83}
    & \metricci{14.54}{8.31}{21.92}
    & \metricci{9.22}{4.79}{14.59}
    & \metricci{1.45}{0.81}{2.30}
    & \metricci{0.49}{0.30}{0.73}
    & \metricci{0.421}{0.111}{0.725}
    & \metricci{0.207}{0.010}{0.509}
    & \metricci{\textbf{0.178}}{0.008}{0.489} \\

    MorphLDM~\citep{wang2025generating}
    & \metricci{46.65}{29.85}{70.74}
    & \metricci{8.56}{5.90}{12.98}
    & \metricci{18.05}{15.54}{21.27}
    & \metricci{2.23}{1.88}{2.62}
    & \metricci{0.99}{0.72}{1.35}
    & \metricci{\textbf{0.131}}{0.005}{0.374}
    & \metricci{\underline{0.202}}{0.007}{0.510}
    & \metricci{0.357}{0.048}{0.662} \\

    \shortstack[l]{ControlNet LDM}
    & \metricci{44.24}{22.72}{71.79}
    & \metricci{8.30}{3.87}{14.15}
    & \metricci{7.58}{4.06}{12.19}
    & \metricci{\underline{1.24}}{0.70}{2.00}
    & \metricci{\underline{0.36}}{0.20}{0.57}
    & \metricci{0.301}{0.022}{0.665}
    & \metricci{1.366}{1.026}{1.761}
    & \metricci{1.311}{0.951}{1.667} \\

    \shortstack[l]{Grid-based LDM}
    & \metricci{64.53}{46.6}{86.0}
    & \metricci{9.72}{7.7}{12.8}
    & \metricci{11.36}{8.6}{15.4}
    & \metricci{3.02}{2.2}{4.5}
    & \metricci{4.83}{3.3}{7.3}
    & \metricci{0.427}{0.082}{0.757}
    & \metricci{1.821}{1.449}{2.265}
    & \metricci{1.235}{0.888}{1.591} \\
    \midrule
    \textbf{Ours}
    & \metricci{\textbf{36.16}}{19.16}{59.61}
    & \metricci{\textbf{6.21}}{3.14}{10.56}
    & \metricci{\textbf{6.67}}{3.67}{10.72}
    & \metricci{\textbf{1.06}}{0.64}{1.64}
    & \metricci{\textbf{0.29}}{0.17}{0.46}
    & \metricci{\underline{0.144}}{0.005}{0.416}
    & \metricci{\textbf{0.169}}{0.007}{0.482}
    & \metricci{\underline{0.185}}{0.008}{0.483} \\
    \bottomrule
  \end{tabular}%
  }
\end{table}

\subsection{Ablation Study}
\label{sec:ablation}
We conduct ablation studies to disentangle the contributions of (a) compositional design, (b) latent intervention strength, (c) injection time variation, and (d) assembly space.

\paragraph{\textbf{Separate Hemisphere Models.}} We ablate the part generator design by using distinct AEs and LDMs for the left and right hemispheres (instead of a shared hemisphere generator) to evaluate whether part-specific latent spaces improve anatomical fidelity. As shown in Table~\ref{tab:ablation_metrics}, separate hemisphere models yield lower mean FID for the left hemisphere, \(5.59\,[2.10,10.47]\) versus \(6.21\,[3.14,10.56]\), the cerebellar--brainstem complex, \(0.89\,[0.43,1.54]\) versus \(1.06\,[0.64,1.64]\), and the seam region, \(0.19\,[0.10,0.35]\) versus \(0.29\,[0.17,0.46]\). However, they yield higher mean FID for the right hemisphere, \(8.94\,[4.29,14.76]\) versus \(6.67\,[3.67,10.72]\), and the whole brain, \(38.72\,[16.83,67.13]\) versus \(36.16\,[19.16,59.61]\). The confidence intervals overlap, so neither design is uniformly better. The shared model still gives lower whole-brain FID (36.16 vs. 38.72) and lower mean bilateral FID (6.44 vs. 7.27), which indicates better bilateral balance in the composed brain. In segmentation, CNNs benefit from explicit spatial context in registered brain volumes and can confuse homologous left/right structures without it \citep{novosad2020accurate}. Canonicalization with side conditioning may likewise share a prior across homologous anatomy, with the side indicator supplying lateral context without enforcing mirror symmetry. We hypothesize a sample-efficiency and regularization trade-off. The shared model pools observations from both sides, while separate models have more side-specific capacity but fewer training examples each and may show more optimization, sampling, or finite-sample FID variability. Direction-dependent effects of representational ordering, reported for sequence models \citep{kutscher2025reordering,hardan2025flatten}, motivate further study but do not directly explain our convolutional U-Net results. We present these explanations as hypotheses, since attributing the asymmetry to biological laterality would require targeted experiments.

\paragraph{\textbf{Latent Injection.}} We evaluate the contribution of latent injection by comparing full scaffold injection (\(\alpha=1\)) with no injection (\(\alpha=0\)). As shown in Table~\ref{tab:ablation_metrics}, disabling injection increases the mean FID across every evaluated region, including the whole brain, from \(36.16\,[19.16,59.61]\) to \(52.27\,[30.69,79.63]\), and the seam region, from \(0.29\,[0.17,0.46]\) to \(0.55\,[0.35,0.81]\). This suggests that injecting the composed part latents provides useful local anatomical priors for regional fidelity and whole-brain composition.

\paragraph{\textbf{Injection Time Variation.}} We vary \(r_{\mathrm{inj}}\) at inference to quantify the trade-off between local part anchoring and global harmonization. Here, \(r_{\mathrm{inj}}\) denotes the number of DDIM steps remaining after part-latent injection. Since training samples injection timesteps from \(t_{\mathrm{inj}}\in[100,300]\) on the 1000-step DDPM schedule, we evaluate \(r_{\mathrm{inj}}\in\{7,10,15\}\), corresponding approximately to \(t_{\mathrm{inj}}\in\{120,180,280\}\). As shown in Table~\ref{tab:ablation_metrics}, \(r_{\mathrm{inj}}=10\) yields the lowest mean FID across all five evaluated regions, although the confidence intervals overlap. This point-estimate pattern is consistent with a balance between retaining enough refinement for seam harmonization and preventing subsequent denoising from weakening the injected part structure.

\paragraph{\textbf{Assembly Space.}} Because the regional and whole-brain autoencoders are trained independently, their latent spaces are not necessarily aligned, even where they have the same dimensionality. AnaDiffusion therefore assembles the decoded part assets in image space and re-encodes the scaffold with the whole-brain encoder. Assembling the part latents without re-encoding raises the mean FID in every region (Table~\ref{tab:ablation_metrics}), from \(36.16\) to \(149.05\) for the whole brain and from \(1.06\) to \(18.23\) for the cerebellar--brainstem complex, and produces intensity shifts and localized structural failures.

\begin{table}[tb]
  \centering
  \small
  \renewcommand{\arraystretch}{1.05}
  \setlength{\tabcolsep}{2.2pt}
  \caption{\looseness=-1 \textbf{Ablation study on \textit{ADNI}.}
We evaluate the effects of compositional design, latent injection strength, injection timing, and assembly space (image-space assembly with re-encoding, as in AnaDiffusion, versus direct assembly of the part latents) on generation quality. 
Values are MedicalNet FID \(\times 10^4\) for the whole brain (WB), left/right hemispheres, cerebellar-brainstem complex (CB), and seam region. 
Best results are \textbf{bold} and second-best are \underline{underlined} (lower is better).}
  \label{tab:ablation_metrics}

  \resizebox{\textwidth}{!}{%
  \begin{tabular}{@{}lccccc@{}}
    \toprule
    \multirow[c]{2}{*}{Method}
    & WB
    & Left Hemi
    & Right Hemi
    & CB
    & Seam \\
    & FID
    & FID
    & FID
    & FID
    & FID \\
    \midrule

    \multicolumn{6}{@{}l}{\textit{Compositional design} \((r_{\mathrm{inj}}=10)\)} \\
    separate hemisphere models
    & \underline{38.72 [16.83, 67.13]}
    & \textbf{5.59 [2.10, 10.47]}
    & 8.94 [4.29, 14.76]
    & \textbf{0.89 [0.43, 1.54]}
    & \textbf{0.19 [0.10, 0.35]} \\
    \midrule

    \multicolumn{6}{@{}l}{\textit{Latent injection strength} \((r_{\mathrm{inj}}=10)\)} \\
    w/o latent injection \((\alpha=0)\)
    & 52.27 [30.69, 79.63]
    & 9.83 [5.40, 15.53]
    & \underline{8.52 [5.23, 12.55]}
    & 1.84 [1.12, 2.75]
    & 0.55 [0.35, 0.81] \\
    full latent injection \((\alpha=1)\) (Ours)
    & \textbf{36.16 [19.16, 59.61]}
    & \underline{6.21 [3.14, 10.56]}
    & \textbf{6.67 [3.67, 10.72]}
    & \underline{1.06 [0.64, 1.64]}
    & \underline{0.29 [0.17, 0.46]} \\
    \midrule

    \multicolumn{6}{@{}l}{\textit{Injection timing}} \\
    \(r_{\mathrm{inj}}=7\)
    & 53.27 [34.21, 76.75]
    & 9.31 [5.88, 13.73]
    & 9.04 [5.88, 12.76]
    & 1.75 [1.22, 2.41]
    & 0.69 [0.51, 0.91] \\
    \(r_{\mathrm{inj}}=15\)
    & 51.66 [28.31, 80.82]
    & 8.96 [4.73, 14.63]
    & 9.52 [5.39, 14.82]
    & 1.43 [0.90, 2.10]
    & 0.31 [0.16, 0.51] \\
    \midrule

    \multicolumn{6}{@{}l}{\textit{Assembly space} \((r_{\mathrm{inj}}=10)\)} \\
    w/o re-encoding (latent assembly)
    & 149.05 [104.63, 199.33]
    & 26.77 [16.99, 37.68]
    & 23.93 [15.95, 32.99]
    & 18.23 [16.20, 20.36]
    & 0.63 [0.45, 0.86] \\
    \bottomrule
  \end{tabular}%
  }
\end{table}

\subsection{Controllable Part Editing and Compositional Refinement}
\label{sec:editing}

We evaluate localized editing using paired multi-scale structural similarity (MS-SSIM) \citep{wang2003multiscale}. For each replacement region, we compare AnaDiffusion with the Seg. cLDM baseline using the same donor-recipient editing pairs. A recipient whole brain is first sampled from the standard LDM, and an independent donor part is sampled from the corresponding part model. We replace the recipient target part with the donor part, keep all other parts fixed, and then assemble, re-encode, inject, and refine the edited asset with the AnaDiffusion whole-brain denoiser. For the Segm. cLDM baseline, we use an SDEdit-style editing protocol \citep{meng2022sdedit} that updates the whole-brain conditioning mask with the selected donor part. We generate \(N=100\) edited scans from 48 subjects per region and report mean MS-SSIM with 95\% confidence intervals from 5,000 subject-cluster bootstrap replicates.

\begin{figure}[t]
\centering
\small

\begin{minipage}[c]{\textwidth}
\centering
\textbf{(a) Quantitative editing locality}\\[0.45em]
\renewcommand{\arraystretch}{1.05}
\setlength{\tabcolsep}{3.0pt}
\resizebox{\linewidth}{!}{%
\begin{tabular}{@{}llcccc@{}}
\toprule
Region & Method & Target transfer $\uparrow$ & Off-target preserv. $\uparrow$ & Transfer gain $\uparrow$ & Locality contrast $\uparrow$ \\
\midrule
\multirow{2}{*}{Left hemisphere}
& Ours & \textbf{0.9427 [0.9411, 0.9441]} & 0.9371 [0.9359, 0.9383] & \textbf{0.1627 [0.1565, 0.1688]} & \textbf{0.1207 [0.1151, 0.1261]} \\
& Segm. cLDM & 0.7952 [0.7887, 0.8018] & \textbf{0.9487 [0.9478, 0.9496]} & 0.0152 [0.0138, 0.0168] & $-0.0078$ [$-0.0090$, $-0.0065$] \\
\addlinespace[0.25em]
\multirow{2}{*}{Right hemisphere}
& Ours & \textbf{0.9484 [0.9471, 0.9495]} & 0.9363 [0.9350, 0.9376] & \textbf{0.1372 [0.1313, 0.1428]} & \textbf{0.1118 [0.1064, 0.1172]} \\
& Segm. cLDM & 0.8343 [0.8300, 0.8384] & \textbf{0.9447 [0.9434, 0.9458]} & 0.0231 [0.0207, 0.0254] & 0.0056 [0.0036, 0.0076] \\
\addlinespace[0.25em]
\multirow{2}{*}{CB complex}
& Ours & \textbf{0.9618 [0.9609, 0.9628]} & 0.9496 [0.9484, 0.9508] & \textbf{0.0752 [0.0716, 0.0788]} & \textbf{0.0313 [0.0278, 0.0348]} \\
& Segm. cLDM & 0.9151 [0.9122, 0.9183] & \textbf{0.9512 [0.9500, 0.9523]} & 0.0285 [0.0269, 0.0302] & $-0.0207$ [$-0.0215$, $-0.0199$] \\
\bottomrule
\end{tabular}%
}
\end{minipage}

\vspace{0.6em}
\begin{minipage}[c]{\textwidth}
\centering
\textbf{(b) Example localized part replacement}\\[0.45em]
\includegraphics[width=\linewidth]{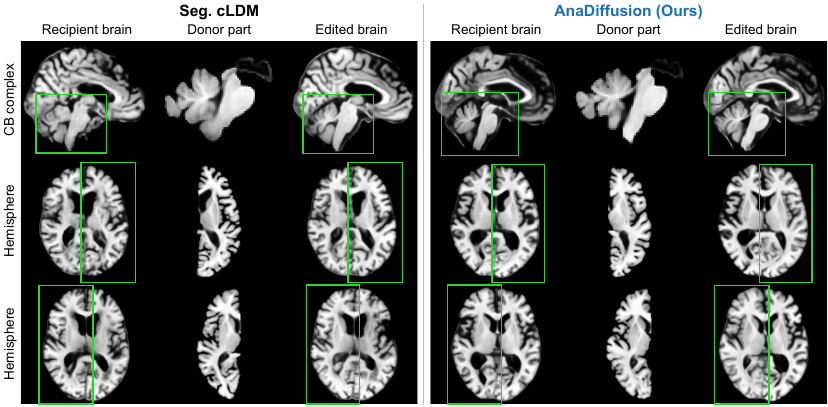}
\end{minipage}

\vspace{-0.35em}
\caption{\textbf{Localized part editing.}
We evaluate whether part replacement edits are both effective and spatially localized.
\textbf{(a)} AnaDiffusion and the SDEdit-style Segm. cLDM baseline are evaluated on the same \(N=100\) donor-recipient edits from 48 subjects per region. Values are mean paired MS-SSIM with 95\% confidence intervals from 5,000 subject-cluster bootstrap replicates.
Target transfer measures similarity between the edited target region and the inserted donor part,
while off-target preservation measures similarity outside the edited region before and after editing.
Contrast metrics further summarize the transfer gain and locality contrast.
\textbf{(b)} A donor part asset is substituted into a generated recipient brain and refined into an edited whole-brain sample.}
\label{fig:editing_locality}
\vspace{-0.6em}
\end{figure}

\looseness=-1 \Cref{fig:editing_locality}a summarizes the quantitative editing results. Target transfer measures whether the inserted donor part is preserved in the final edited brain, while off-target preservation measures whether the non-edited anatomy remains unchanged. AnaDiffusion achieves substantially higher target transfer than Segm. cLDM for the left hemisphere (0.9427 vs. 0.7952), right hemisphere (0.9484 vs. 0.8343), and cerebellar--brainstem complex (0.9618 vs. 0.9151). Off-target preservation remains consistently high for AnaDiffusion (0.9363--0.9496), although it is slightly lower than for Segm. cLDM (0.9447--0.9512). AnaDiffusion also yields larger transfer gains and positive locality contrasts across all three regions, indicating that the edits move the target anatomy toward the donor while concentrating changes within the intended region. The qualitative example in \Cref{fig:editing_locality}b further shows that the substituted part can be inserted into the recipient brain without disrupting the surrounding anatomy.

\section{Conclusion}
\label{sec:conclusion}

In this work, we presented \modelname, a part-to-whole latent diffusion framework for controllable 3D brain MRI generation. 
The central motivation behind \modelname is that brain MRI synthesis should not be treated purely as monolithic volume generation: anatomical regions have distinct local distributions, interact through structured interfaces, and often require region-specific control. 
By explicitly generating anatomical part assets and refining their assembled scaffold with a whole-brain denoiser, \modelname introduces anatomical compositionality into latent diffusion generation. The output is a coherent whole-brain sample with reusable, editable anatomical components. 
\modelname attains the lowest mean FID in every evaluated region and clearly higher target transfer in localized editing than a segmentation-conditioned baseline, without subject-specific dense segmentation masks at inference; ablations suggest that latent injection, the injection schedule, and image-space assembly contribute to these results. 

\paragraph{\textbf{Limitations.}}
Despite these advantages, several limitations remain. 
First, the multi-stage design introduces additional training and inference complexity compared with a monolithic LDM. 
Second, the current fixed, anatomy-driven factorization improves local part and seam behavior, but does not fully solve broader tissue-level calibration. 
Third, the chosen decomposition focuses on predefined anatomical regions and may not capture other biologically meaningful organizations, such as tissue classes, functional networks, or multi-scale anatomical hierarchies. Fourth, AnaDiffusion assumes approximate correspondence with MNI152 and has not been validated for severe mass effect or displaced anatomical boundaries. Such pathology may impair registration and invalidate fixed regional placements. Supporting these cases would require lesion-aware or subject-adaptive localization.

\paragraph{\textbf{Future work.}}
Future work will explore learned, adaptive decompositions beyond the three coarse parts, such as finer subcortical or lobar units, and will compare them directly with anatomy-free factorizations. Assembly could move into latent space through a shared autoencoder or a learned cross-latent mapping, which would remove the decode and re-encode step and reduce the cost of the multi-stage pipeline. Region-dependent refinement schedules may let the whole-brain denoiser spend more steps on parts and seams with complex morphology. The modular design also offers a path toward condition-specific synthesis. A selected part generator could be trained or fine-tuned with a clinical or morphological condition, such as age, diagnosis, or localized atrophy, and its output inserted and reconciled through whole-brain refinement. Finally, we will train segmentation models on generated volumes and evaluate the approach on other cohorts, scanners, pathologies, resolutions, and imaging contrasts, including younger populations.

\section{Acknowledgements}

This work was supported in part by GPU hardware and cloud credits provided to the Texas A\&M University
PHAI Lab through the NVIDIA Academic Grant Program. Portions of this research were conducted with the advanced computing resources provided by Texas A\&M High Performance Research Computing, including VISION GPU Cluster. Research reported in this publication was supported by the National Institute On Aging of the National Institutes of Health under Award Numbers U01AG070112 and R01AG081398. The content is solely the responsibility of the authors and does not necessarily represent the official views of the National Institutes of Health. The authors declare no competing interests.

\begingroup
\renewcommand{\thefootnote}{\fnsymbol{footnote}}
Data collection and sharing for the Alzheimer's Disease Neuroimaging Initiative\footnote[1]{Data used in preparation of this article were obtained from the Alzheimer's Disease Neuroimaging Initiative (ADNI) database (\url{adni.loni.usc.edu}). As such, the investigators within the ADNI contributed to the design and implementation of ADNI and/or provided data but did not participate in analysis or writing of this report. A complete listing of ADNI investigators can be found at: \url{http://adni.loni.usc.edu/wp-content/uploads/how_to_apply/ADNI_Acknowledgement_List.pdf}} (ADNI) is funded by the National Institute on Aging (National Institutes of Health Grant U19 AG024904).
The grantee organization is the Northern California Institute for Research and Education.
In the past, ADNI has also received funding from the National Institute of Biomedical Imaging and Bioengineering, the Canadian Institutes of Health Research, and private sector contributions through the Foundation for the National Institutes of Health (FNIH) including generous contributions from the following: AbbVie, Alzheimer's Association; Alzheimer's Drug Discovery Foundation; Araclon Biotech; BioClinica, Inc.; Biogen; Bristol-Myers Squibb Company; CereSpir, Inc.; Cogstate; Eisai Inc.; Elan Pharmaceuticals, Inc.; Eli Lilly and Company; EuroImmun; F. Hoffmann-La Roche Ltd and its affiliated company Genentech, Inc.; Fujirebio; GE Healthcare; IXICO Ltd.; Janssen Alzheimer Immunotherapy Research \& Development, LLC.; Johnson \& Johnson Pharmaceutical Research \& Development LLC.; Lumosity; Lundbeck; Merck \& Co., Inc.; Meso Scale Diagnostics, LLC.; NeuroRx Research; Neurotrack Technologies; Novartis Pharmaceuticals Corporation; Pfizer Inc.; Piramal Imaging; Servier; Takeda Pharmaceutical Company; and Transition Therapeutics.
\par
\endgroup

\clearpage
\bibliographystyle{assets/plainnat}
\bibliography{main}

@inproceedings{peng2023generating,
  title={Generating Realistic Brain {MRIs} via a Conditional Diffusion Probabilistic Model},
  author={Peng, Wei and Adeli, Ehsan and Bosschieter, Tomas and Park, Sang Hyun and Zhao, Qingyu and Pohl, Kilian M},
  booktitle={International conference on medical image computing and computer-assisted intervention},
  pages={14--24},
  year={2023},
  organization={Springer}
}

@article{dorjsembe2024conditional,
  title={Conditional diffusion models for semantic {3D} brain {MRI} synthesis},
  author={Dorjsembe, Zolnamar and Pao, Hsing-Kuo and Odonchimed, Sodtavilan and Xiao, Furen},
  journal={IEEE Journal of Biomedical and Health Informatics},
  volume={28},
  number={7},
  pages={4084--4093},
  year={2024},
  publisher={IEEE}
}

@inproceedings{wang2025generating,
  title={Generating novel brain morphology by deforming learned templates},
  author={Wang, Alan Q and Huang, Fangrui and Trang, Bailey and Peng, Wei and Abbasi, Mohammad and Pohl, Kilian and Sabuncu, Mert R and Adeli, Ehsan},
  booktitle={International Conference on Medical Image Computing and Computer-Assisted Intervention},
  pages={207--217},
  year={2025},
  organization={Springer}
}

@article{bongratz2026cortex,
  title={Cortex-Grounded Diffusion Models for Brain Image Generation},
  author={Bongratz, Fabian and Li, Yitong and Elbaroudy, Sama and Wachinger, Christian},
  journal={arXiv preprint arXiv:2601.19498},
  year={2026}
}

@article{billot2023synthseg,
  title={{SynthSeg}: Segmentation of brain {MRI} scans of any contrast and resolution without retraining},
  author={Billot, Benjamin and Greve, Douglas N and Puonti, Oula and Thielscher, Axel and Van Leemput, Koen and Fischl, Bruce and Dalca, Adrian V and Iglesias, Juan Eugenio and others},
  journal={Medical image analysis},
  volume={86},
  pages={102789},
  year={2023},
  publisher={Elsevier}
}

@inproceedings{puglisi2024enhancing,
  title={Enhancing spatiotemporal disease progression models via latent diffusion and prior knowledge},
  author={Puglisi, Lemuel and Alexander, Daniel C and Rav{\`\i}, Daniele},
  booktitle={International Conference on Medical Image Computing and Computer-Assisted Intervention},
  pages={173--183},
  year={2024},
  organization={Springer}
}

@inproceedings{konz2024anatomically,
  title={Anatomically-controllable medical image generation with segmentation-guided diffusion models},
  author={Konz, Nicholas and Chen, Yuwen and Dong, Haoyu and Mazurowski, Maciej A},
  booktitle={International Conference on Medical Image Computing and Computer-Assisted Intervention},
  pages={88--98},
  year={2024},
  organization={Springer}
}

@inproceedings{dhinagar2024counterfactual,
  title={Counterfactual {MRI} generation with denoising diffusion models for interpretable {Alzheimer's} disease effect detection},
  author={Dhinagar, Nikhil J and Thomopoulos, Sophia I and Laltoo, Emily and Thompson, Paul M},
  booktitle={2024 46th Annual International Conference of the IEEE Engineering in Medicine and Biology Society (EMBC)},
  pages={1--6},
  year={2024},
  organization={IEEE}
}

@article{fu2025synthesizing,
  title={Synthesizing individualized aging brains in health and disease with generative models and parallel transport},
  author={Fu, Jingru and Zheng, Yuqi and Dey, Neel and Ferreira, Daniel and Moreno, Rodrigo},
  journal={Medical Image Analysis},
  volume={105},
  pages={103669},
  year={2025},
  publisher={Elsevier}
}

@inproceedings{kim2024adaptive,
  title={Adaptive latent diffusion model for {3D} medical image to image translation: Multi-modal magnetic resonance imaging study},
  author={Kim, Jonghun and Park, Hyunjin},
  booktitle={Proceedings of the IEEE/CVF Winter conference on applications of computer Vision},
  pages={7604--7613},
  year={2024}
}

@inproceedings{wang2023inversesr,
  title={{InverseSR}: {3D} Brain {MRI} Super-Resolution Using a Latent Diffusion Model},
  author={Wang, Jueqi and Levman, Jacob and Pinaya, Walter Hugo Lopez and Tudosiu, Petru-Daniel and Cardoso, M Jorge and Marinescu, Razvan},
  booktitle={International conference on medical image computing and computer-assisted intervention},
  pages={438--447},
  year={2023},
  organization={Springer}
}

@inproceedings{pinaya2022brain,
  title={Brain imaging generation with latent diffusion models},
  author={Pinaya, Walter HL and Tudosiu, Petru-Daniel and Dafflon, Jessica and Da Costa, Pedro F and Fernandez, Virginia and Nachev, Parashkev and Ourselin, Sebastien and Cardoso, M Jorge},
  booktitle={MICCAI workshop on deep generative models},
  pages={117--126},
  year={2022},
  organization={Springer}
}

@article{bhattacharya2025brainmrdiff,
  title={{BrainMRDiff}: A Diffusion Model for Anatomically Consistent Brain {MRI} Synthesis},
  author={Bhattacharya, Moinak and Gupta, Saumya and Singh, Annie and Chen, Chao and Singh, Gagandeep and Prasanna, Prateek},
  journal={arXiv preprint arXiv:2504.04532},
  year={2025}
}

@article{khateri2025mri,
  title={{MRI} Super-Resolution with Deep Learning: A Comprehensive Survey},
  author={Khateri, Mohammad and Vasylechko, Serge and Ghahremani, Morteza and Timms, Liam and Kocanaogullari, Deniz and Warfield, Simon K and Jaimes, Camilo and Karimi, Davood and Sierra, Alejandra and Tohka, Jussi and others},
  journal={arXiv preprint arXiv:2511.16854},
  year={2025}
}

@inproceedings{wu2024evaluating,
  title={Evaluating the quality of brain {MRI} generators},
  author={Wu, Jiaqi and Peng, Wei and Li, Binxu and Zhang, Yu and Pohl, Kilian M},
  booktitle={Medical Image Computing and Computer Assisted Intervention -- MICCAI 2024},
  volume={15010},
  pages={297--307},
  year={2024},
  publisher={Springer Nature Switzerland},
  doi={10.1007/978-3-031-72117-5_28}
}

@inproceedings{kutscher2025reordering,
  title={{REOrdering Patches Improves Vision Models}},
  author={Kutscher, Declan and Chan, David M. and Bai, Yutong and Darrell, Trevor and Gupta, Ritwik},
  booktitle={Advances in Neural Information Processing Systems},
  year={2025},
  url={https://openreview.net/forum?id=g56WiaXKGF}
}

@article{hardan2025flatten,
  title={Flatten Wisely: How Patch Order Shapes {Mamba}-Powered Vision for {MRI} Segmentation},
  author={Hardan, Osama and Elshenhabi, Omar and Khattab, Tamer and Mabrok, Mohamed},
  journal={arXiv preprint arXiv:2507.13384},
  year={2025},
  doi={10.48550/arXiv.2507.13384}
}

@article{taha2015metrics,
  title={Metrics for evaluating {3D} medical image segmentation: analysis, selection, and tool},
  author={Taha, Abdel Aziz and Hanbury, Allan},
  journal={BMC Medical Imaging},
  volume={15},
  number={1},
  pages={29},
  year={2015},
  doi={10.1186/s12880-015-0068-x}
}

@article{deo2025metrics,
  title={Metrics that matter: Evaluating image quality metrics for medical image generation},
  author={Deo, Yash and Jia, Yan and Lassila, Toni and Smith, William AP and Lawton, Tom and Kang, Siyuan and Frangi, Alejandro F and Habli, Ibrahim},
  journal={arXiv preprint arXiv:2505.07175},
  year={2025}
}

@article{wan2026anatomically,
  title={Anatomically Guided Latent Diffusion for Brain {MRI} Progression Modeling},
  author={Wan, Cheng and Jafrasteh, Bahram and Adeli, Ehsan and Zhang, Miaomiao and Zhao, Qingyu},
  journal={arXiv preprint arXiv:2601.14584},
  year={2026}
}

@article{fernandez2024generating,
  title={Generating multi-pathological and multi-modal images and labels for brain {MRI}},
  author={Fernandez, Virginia and Pinaya, Walter Hugo Lopez and Borges, Pedro and Graham, Mark S and Tudosiu, Petru-Daniel and Vercauteren, Tom and Cardoso, M Jorge},
  journal={Medical Image Analysis},
  volume={97},
  pages={103278},
  year={2024},
  publisher={Elsevier}
}

@inproceedings{jafrasteh2025wasabi,
  title={{WASABI}: A Metric for Evaluating Morphometric Plausibility of Synthetic Brain {MRIs}},
  author={Jafrasteh, Bahram and Peng, Wei and Wan, Cheng and Luo, Yimin and Adeli, Ehsan and Zhao, Qingyu},
  booktitle={International Conference on Medical Image Computing and Computer-Assisted Intervention},
  pages={684--694},
  year={2025},
  organization={Springer}
}

@inproceedings{wu2025igg,
  title={{IGG}: Image Generation Informed by Geodesic Dynamics in Deformation Spaces},
  author={Wu, Nian and Jayakumar, Nivetha and Xing, Jiarui and Zhang, Miaomiao},
  booktitle={International Conference on Information Processing in Medical Imaging},
  pages={232--246},
  year={2025},
  organization={Springer}
}

@article{wilms2022invertible,
  title={Invertible modeling of bidirectional relationships in neuroimaging with normalizing flows: application to brain aging},
  author={Wilms, Matthias and Bannister, Jordan J and Mouches, Pauline and MacDonald, M Ethan and Rajashekar, Deepthi and Langner, S{\"o}nke and Forkert, Nils D},
  journal={IEEE Transactions on Medical Imaging},
  volume={41},
  number={9},
  pages={2331--2347},
  year={2022},
  publisher={IEEE}
}

@article{rusak2022quantifiable,
  title={Quantifiable brain atrophy synthesis for benchmarking of cortical thickness estimation methods},
  author={Rusak, Filip and Santa Cruz, Rodrigo and Lebrat, L{\'e}o and Hlinka, Ondrej and Fripp, Jurgen and Smith, Elliot and Fookes, Clinton and Bradley, Andrew P and Bourgeat, Pierrick and Alzheimer's Disease Neuroimaging Initiative and others},
  journal={Medical Image Analysis},
  volume={82},
  pages={102576},
  year={2022},
  publisher={Elsevier}
}

@article{fischl2002whole,
  title={Whole brain segmentation: automated labeling of neuroanatomical structures in the human brain},
  author={Fischl, Bruce and Salat, David H and Busa, Evelina and Albert, Marilyn and Dieterich, Megan and Haselgrove, Christian and Van Der Kouwe, Andre and Killiany, Ron and Kennedy, David and Klaveness, Shuna and others},
  journal={Neuron},
  volume={33},
  number={3},
  pages={341--355},
  year={2002},
  publisher={Elsevier}
}

@article{ho2020denoising,
  title={Denoising diffusion probabilistic models},
  author={Ho, Jonathan and Jain, Ajay and Abbeel, Pieter},
  journal={Advances in neural information processing systems},
  volume={33},
  pages={6840--6851},
  year={2020}
}

@inproceedings{rombach2022high,
  title={High-resolution image synthesis with latent diffusion models},
  author={Rombach, Robin and Blattmann, Andreas and Lorenz, Dominik and Esser, Patrick and Ommer, Bj{\"o}rn},
  booktitle={Proceedings of the IEEE/CVF conference on computer vision and pattern recognition},
  pages={10684--10695},
  year={2022}
}

@inproceedings{kwon2019generation,
  title={Generation of {3D} brain {MRI} using auto-encoding generative adversarial networks},
  author={Kwon, Gihyun and Han, Chihye and Kim, Dae-shik},
  booktitle={International Conference on Medical Image Computing and Computer-Assisted Intervention},
  pages={118--126},
  year={2019},
  organization={Springer}
}

@inproceedings{xing2021cycle,
  title={Cycle consistent embedding of {3D} brains with auto-encoding generative adversarial networks},
  author={Xing, Shibo and Sinha, Harsh and Hwang, Seong Jae},
  booktitle={Medical Imaging with Deep Learning},
  year={2021}
}

@inproceedings{dorjsembe2022three,
  title={Three-dimensional medical image synthesis with denoising diffusion probabilistic models},
  author={Dorjsembe, Zolnamar and Odonchimed, Sodtavilan and Xiao, Furen},
  booktitle={Medical imaging with deep learning},
  year={2022}
}

@inproceedings{meng2022sdedit,
  title={{SDEdit}: Guided image synthesis and editing with stochastic differential equations},
  author={Meng, Chenlin and He, Yutong and Song, Yang and Song, Jiaming and Wu, Jiajun and Zhu, Jun-Yan and Ermon, Stefano},
  booktitle={International Conference on Learning Representations},
  year={2022},
  url={https://openreview.net/forum?id=aBsCjcPu_tE}
}

@article{tustison2010n4itk,
  title={{N4ITK}: improved {N3} bias correction},
  author={Tustison, Nicholas J and Avants, Brian B and Cook, Philip A and Zheng, Yuanjie and Egan, Alexander and Yushkevich, Paul A and Gee, James C},
  journal={IEEE transactions on medical imaging},
  volume={29},
  number={6},
  pages={1310--1320},
  year={2010},
  publisher={IEEE}
}

@article{avants2008symmetric,
  title={Symmetric diffeomorphic image registration with cross-correlation: evaluating automated labeling of elderly and neurodegenerative brain},
  author={Avants, Brian B and Epstein, Charles L and Grossman, Murray and Gee, James C},
  journal={Medical image analysis},
  volume={12},
  number={1},
  pages={26--41},
  year={2008},
  publisher={Elsevier}
}

@article{shinohara2014statistical,
  title={Statistical normalization techniques for magnetic resonance imaging},
  author={Shinohara, Russell T and Sweeney, Elizabeth M and Goldsmith, Jeff and Shiee, Navid and Mateen, Farrah J and Calabresi, Peter A and Jarso, Samson and Pham, Dzung L and Reich, Daniel S and Crainiceanu, Ciprian M and others},
  journal={NeuroImage: Clinical},
  volume={6},
  pages={9--19},
  year={2014},
  publisher={Elsevier}
}

@article{hoopes2022synthstrip,
  title={{SynthStrip}: skull-stripping for any brain image},
  author={Hoopes, Andrew and Mora, Jocelyn S and Dalca, Adrian V and Fischl, Bruce and Hoffmann, Malte},
  journal={NeuroImage},
  volume={260},
  pages={119474},
  year={2022},
  publisher={Elsevier}
}

@article{iglesias2023ready,
  title={A ready-to-use machine learning tool for symmetric multi-modality registration of brain {MRI}},
  author={Iglesias, Juan Eugenio},
  journal={Scientific Reports},
  volume={13},
  number={1},
  pages={6657},
  year={2023},
  publisher={Nature Publishing Group UK London}
}

@article{gui2012morphology,
  title={Morphology-driven automatic segmentation of {MR} images of the neonatal brain},
  author={Gui, Laura and Lisowski, Radoslaw and Faundez, Tamara and H{\"u}ppi, Petra S and Lazeyras, Fran{\c{c}}ois and Kocher, Michel},
  journal={Medical image analysis},
  volume={16},
  number={8},
  pages={1565--1579},
  year={2012},
  publisher={Elsevier}
}

@article{liang2007automatic,
  title={Automatic segmentation of left and right cerebral hemispheres from {MRI} brain volumes using the graph cuts algorithm},
  author={Liang, Lichen and Rehm, Kelly and Woods, Roger P and Rottenberg, David A},
  journal={NeuroImage},
  volume={34},
  number={3},
  pages={1160--1170},
  year={2007},
  publisher={Elsevier}
}

@article{cardoso2022monai,
  title={{MONAI}: An Open-Source Framework for Deep Learning in Healthcare},
  author={Cardoso, M Jorge and Li, Wenqi and Brown, Richard and Ma, Nic and Kerfoot, Eric and Wang, Yiheng and Murrey, Benjamin and Myronenko, Andriy and Zhao, Can and Yang, Dong and others},
  journal={arXiv preprint arXiv:2211.02701},
  year={2022}
}

@article{pinaya2023generative,
  title={Generative {AI} for Medical Imaging: Extending the {MONAI} Framework},
  author={Pinaya, Walter HL and Graham, Mark S and Kerfoot, Eric and Tudosiu, Petru-Daniel and Dafflon, Jessica and Fernandez, Virginia and Sanchez, Pedro and Wolleb, Julia and Da Costa, Pedro F and Patel, Ashay and others},
  journal={arXiv preprint arXiv:2307.15208},
  year={2023}
}

@article{paszke2019pytorch,
  title={{PyTorch}: An Imperative Style, High-Performance Deep Learning Library},
  author={Paszke, Adam and Gross, Sam and Massa, Francisco and Lerer, Adam and Bradbury, James and Chanan, Gregory and Killeen, Trevor and Lin, Zeming and Gimelshein, Natalia and Antiga, Luca and others},
  journal={Advances in neural information processing systems},
  volume={32},
  year={2019}
}

@misc{guide2013cuda,
  author={{NVIDIA Corporation}},
  title={{CUDA C++} Programming Guide},
  year={2025},
  note={Version 12.9},
  howpublished={\url{https://docs.nvidia.com/cuda/archive/12.9.1/cuda-c-programming-guide/index.html}}
}

@article{saad2024survey,
  title={A survey on training challenges in generative adversarial networks for biomedical image analysis},
  author={Saad, Muhammad Muneeb and O'Reilly, Ruairi and Rehmani, Mubashir Husain},
  journal={Artificial Intelligence Review},
  volume={57},
  number={2},
  pages={19},
  year={2024},
  publisher={Springer}
}

@article{rosca2017variational,
  title={Variational approaches for auto-encoding generative adversarial networks},
  author={Rosca, Mihaela and Lakshminarayanan, Balaji and Warde-Farley, David and Mohamed, Shakir},
  journal={arXiv preprint arXiv:1706.04987},
  year={2017}
}

@inproceedings{wang2003multiscale,
  title={Multiscale structural similarity for image quality assessment},
  author={Wang, Zhou and Simoncelli, Eero P and Bovik, Alan C},
  booktitle={The Thirty-Seventh Asilomar Conference on Signals, Systems \& Computers, 2003},
  volume={2},
  pages={1398--1402},
  year={2003},
  organization={IEEE}
}

@inproceedings{song2020denoising,
  title={Denoising Diffusion Implicit Models},
  author={Song, Jiaming and Meng, Chenlin and Ermon, Stefano},
  booktitle={International Conference on Learning Representations},
  year={2021}
}

@article{herencia2025diffusion,
  title={Diffusion Models for Conditional {MRI} Generation},
  author={Herencia Garc{\'\i}a del Castillo, Miguel and Moya Garcia, Ricardo and Cerezo Maz{\'o}n, Manuel Jes{\'u}s and Arriola Garcia, Ekaitz and Men{\'e}ndez Fern{\'a}ndez-Miranda, Pablo},
  journal={arXiv preprint arXiv:2502.18620},
  year={2025},
  url={https://arxiv.org/abs/2502.18620}
}

@article{sun2022hierarchical,
  title={Hierarchical amortized {GAN} for {3D} high resolution medical image synthesis},
  author={Sun, Li and Chen, Junxiang and Xu, Yanwu and Gong, Mingming and Yu, Ke and Batmanghelich, Kayhan},
  journal={IEEE journal of biomedical and health informatics},
  volume={26},
  number={8},
  pages={3966--3975},
  year={2022},
  publisher={IEEE}
}

@article{mazziotta1995probabilistic,
  title={A probabilistic atlas of the human brain: Theory and rationale for its development: The international consortium for brain mapping (icbm)},
  author={Mazziotta, John C and Toga, Arthur W and Evans, Alan and Fox, Peter and Lancaster, Jack},
  journal={Neuroimage},
  volume={2},
  number={2},
  pages={89--101},
  year={1995},
  publisher={Elsevier}
}

@article{chen2019med3d,
    title={{Med3D}: Transfer Learning for {3D} Medical Image Analysis},
    author={Chen, Sihong and Ma, Kai and Zheng, Yefeng},
    journal={arXiv preprint arXiv:1904.00625},
    year={2019}
}

@article{novosad2020accurate,
  title={Accurate and robust segmentation of neuroanatomy in {T1-weighted} {MRI} by combining spatial priors with deep convolutional neural networks},
  author={Novosad, Philip and Fonov, Vladimir and Collins, D Louis and {Alzheimer's Disease Neuroimaging Initiative}},
  journal={Human brain mapping},
  volume={41},
  number={2},
  pages={309--327},
  year={2020},
  publisher={Wiley Online Library}
}

@inproceedings{larsen2016autoencoding,
  title={Autoencoding beyond pixels using a learned similarity metric},
  author={Larsen, Anders Boesen Lindbo and S{\o}nderby, S{\o}ren Kaae and Larochelle, Hugo and Winther, Ole},
  booktitle={Proceedings of the 33rd International Conference on Machine Learning},
  series={Proceedings of Machine Learning Research},
  volume={48},
  pages={1558--1566},
  year={2016}
}

@article{huang2026neurojepa,
  title={Learning Sparse Latent Predictive Foundation Model for Multimodal Neuroimaging},
  author={Huang, Haoxu and Chen, Long and Chen, Jingyun and Hyun, Jinu and Loftus, James Ryan and Melmed, Kara and Orringer, Daniel and Frontera, Jennifer and Dehkharghani, Seena and Masurkar, Arjun and Razavian, Narges},
  journal={arXiv preprint arXiv:2606.14957},
  year={2026}
}

\clearpage
\beginappendix

\section{Technical Appendices and Supplementary Material}
\label{sec:supp_parts}

\subsection{Training and Inference}
\label{sec:supp_training}
We adopt the AutoencoderKL and diffusion UNet architectures from the MONAI Generative Models library \citep{pinaya2023generative}, built on MONAI \citep{cardoso2022monai}. All models are implemented in PyTorch \citep{paszke2019pytorch} with CUDA \citep{guide2013cuda} and trained with the AdamW optimizer. The final generation resolution is $128\times128\times128$ for the whole brain, $64\times128\times128$ for each hemisphere, and $128\times96\times64$ for the cerebellar-brainstem complex. For autoencoder training, we use a learning rate of $1\times10^{-4}$ for the whole-brain model and $5\times10^{-5}$ for part models. We warm-start training for 20 epochs before enabling adversarial supervision using a patch-based discriminator, with the discriminator learning rate set to half of the corresponding autoencoder learning rate. Autoencoders are trained for up to 100 epochs, and we select the checkpoint with the lowest exponential moving average (EMA) of validation $L1$ reconstruction loss. For diffusion training, all autoencoders are frozen. All diffusion UNet models are trained for 140,000 optimizer steps. We use a 1,000-step DDPM \citep{ho2020denoising} process with a scaled-linear $\beta$ schedule ($\beta_{\text{start}}=0.0015$, $\beta_{\text{end}}=0.0195$) and DDIM \citep{song2020denoising} sampling with 50 steps at inference. For injecting part latents, we sample $t_{\mathrm{inj}}\in[100,300]$ during training to balance (i) having sufficient global structure in the noisy latent for contextual cues and (ii) retaining enough remaining denoising steps for the injected constraint to propagate. Injecting substantially earlier can weaken part identity, while injecting too late provides insufficient propagation. At inference,
we inject with \(r_{\mathrm{inj}}=10\) DDIM steps remaining, which corresponds to
approximately \(t_{\mathrm{inj}}\approx180\) under the 50-step DDIM subsampling.
This matches the training noise regime while retaining enough reverse steps for
part propagation and seam harmonization. All models were trained on H200 GPUs. Part refinement uses latent-space SDEdit with 100 DDIM steps. During training, the part models start from a timestep five times the injection timestep, capped at \(T=1000\) (500--999 for injection timesteps in \([100,300]\)); at inference, they start from the injection timestep. The stronger training perturbation exposes the whole-brain denoiser to large part--scaffold misalignments and keeps the parts from reproducing the whole-brain model's own output. The seam loss uses a band of two latent voxels on each side of the part boundaries, a weight of 0.5, and the noise-level weighting of \Cref{sec:assemble_refine}. FID features come from the MedicalNet ResNet-10 weights released on PyTorch Hub (\texttt{Warvito/MedicalNet-models}), with intensities z-scored within each region; for evaluation, the seam band has a half-width of 2 voxels.

\subsection{Parts Factorization and Assembly Details}
\label{sec:supp_assembly}
\paragraph{Part Definition.}
All structural MRIs are first registered to the common MNI152 space \citep{avants2008symmetric, iglesias2023ready, mazziotta1995probabilistic},  after which anatomical labels are obtained using SynthSeg \citep{billot2023synthseg}. Based on these label maps, we define three fixed anatomical parts for compositional generation: the \emph{left hemisphere}, the \emph{right hemisphere}, and the \emph{cerebellar-brainstem complex}. Specifically, the left hemisphere is formed from SynthSeg labels \(\{2,3,4,5,10,11,12,13,17,18,26,28\}\), the right hemisphere from \(\{41,42,43,\\44,49,50,51,52,53,54,58,60\}\), and the cerebellar-brainstem complex from \(\{7,8,\\46,47,14,15,16\}\). The only explicit exclusion is CSF (label 24), which is not assigned to any part mask and is instead synthesized during the subsequent whole-brain refinement stage. 

\paragraph{Part Cropping.}
\looseness=-1 Registration to the common MNI152 space allows us to use deterministic, template-aligned crops that generalize consistently across subjects. Within the \(128\times128\times128\) whole-brain grid, the left and right hemisphere crops (\(64\times128\times128\)) are obtained by splitting the volume along the left-right axis (midline of the brain), while the cerebellar-brainstem crop (\(128\times96\times64\)) is taken from the inferior posterior portion of the registered brain. Because all samples share the same canonical coordinate system, these fixed crops can be mapped back unambiguously to their original locations when pasting refined parts into the whole-brain canvas during assembly to compose \(\tilde{x}^{\mathrm{scaf}}\).

\paragraph{Part Masks.}
\looseness=-1 To obtain spatially consistent template masks, we segment the MNI152 template using SynthSeg and merge labels according to the three part definitions above, yielding one binary mask for the left hemisphere, one for the right hemisphere, and one for the cerebellar--brainstem complex. Each mask is dilated by 2 voxels (\(3\,\mathrm{mm}\) at \(1.5\,\mathrm{mm}\) isotropic resolution) to provide a small anatomical margin and to accommodate residual inter-subject variation after registration. These dilated masks, together with fixed bounding boxes, define the crops passed to the part models. Refined parts are pasted with the undilated masks intersected with the brain envelope; the cerebellar--brainstem mask is additionally eroded by 4 voxels, and each paste mask is feathered with a Gaussian of standard deviation 1.5 voxels. Inference needs no subject-specific segmentation or mask.

\subsection{Structural Agreement with Real Test Volumes}
\label{sec:supp_structural_agreement}
\looseness=-1 To complement the distributional metrics, we compare SynthSeg-derived label maps of generated and real held-out test volumes for 17 anatomical regions. Dice measures volumetric overlap and HD95 the symmetric 95th-percentile Hausdorff surface distance in millimetres~\citep{taha2015metrics}. Compared with the LDM baseline, AnaDiffusion achieves higher Dice in 16 of 17 regions and lower HD95 in 16 of 17 regions (Table~\ref{tab:structural_agreement}); the exceptions are the cerebral cortex for Dice and CSF for HD95. SynthSeg's training subjects are not public, so we cannot rule out that some held-out ADNI subjects contributed to its training labels. This applies to all SynthSeg-based results, including \(|d|\) in \Cref{tab:main_metrics}.

\begin{table}[h]
  \centering
  \small
  \caption{\textbf{Structural agreement between generated and real test volumes} using SynthSeg-derived anatomical labels. Dice (\(\uparrow\)) and HD95 in mm (\(\downarrow\)); the better value of each pair is in \textbf{bold}.}
  \label{tab:structural_agreement}
  \begin{tabular}{@{}lcccc@{}}
    \toprule
    & \multicolumn{2}{c}{LDM} & \multicolumn{2}{c}{Ours} \\
    \cmidrule(lr){2-3} \cmidrule(l){4-5}
    Region & Dice & HD95 & Dice & HD95 \\
    \midrule
    Cerebral white matter & 0.6461 & 3.88 & \textbf{0.6657} & \textbf{3.61} \\
    Cerebral cortex & \textbf{0.5475} & 2.98 & 0.5353 & \textbf{2.68} \\
    Lateral ventricles & 0.6941 & 5.88 & \textbf{0.8127} & \textbf{3.44} \\
    Inferior lateral ventricles & 0.3257 & 4.45 & \textbf{0.4497} & \textbf{3.37} \\
    Cerebellar white matter & 0.6764 & 3.95 & \textbf{0.7516} & \textbf{3.03} \\
    Cerebellar cortex & 0.7457 & 4.01 & \textbf{0.8031} & \textbf{2.79} \\
    Thalamus & 0.7500 & 3.49 & \textbf{0.8245} & \textbf{2.49} \\
    Caudate & 0.5341 & 3.87 & \textbf{0.6984} & \textbf{2.80} \\
    Putamen & 0.7043 & 3.10 & \textbf{0.7868} & \textbf{2.25} \\
    Pallidum & 0.6165 & 3.28 & \textbf{0.7460} & \textbf{2.21} \\
    Third ventricle & 0.6136 & 3.42 & \textbf{0.7323} & \textbf{2.30} \\
    Fourth ventricle & 0.4588 & 3.83 & \textbf{0.6524} & \textbf{2.52} \\
    Hippocampus & 0.6220 & 3.40 & \textbf{0.7044} & \textbf{2.59} \\
    Amygdala & 0.6322 & 3.20 & \textbf{0.7161} & \textbf{2.54} \\
    CSF & 0.4162 & \textbf{4.10} & \textbf{0.4326} & 4.25 \\
    Nucleus accumbens & 0.4856 & 3.15 & \textbf{0.6296} & \textbf{2.20} \\
    Ventral diencephalon & 0.6605 & 3.05 & \textbf{0.7595} & \textbf{2.19} \\
    \bottomrule
  \end{tabular}
\end{table}

\subsection{ADNI Data}
\label{sec:supp_adni}
Data used in the preparation of this article were obtained from the Alzheimer's Disease Neuroimaging Initiative (ADNI) database (\href{https://adni.loni.usc.edu}{adni.loni.usc.edu}). The ADNI was launched in 2003 as a public-private partnership, led by Principal Investigator Michael W. Weiner, MD. The primary goal of ADNI has been to test whether serial magnetic resonance imaging (MRI), positron emission tomography (PET), other biological markers, and clinical and neuropsychological assessment can be combined to measure the progression of mild cognitive impairment (MCI) and early Alzheimer's disease (AD).

\subsection{Broader Impacts}
\label{sec:supp_broader}
Synthetic brain MRIs with controllable regional anatomy could support data augmentation for under-represented anatomical phenotypes, counterfactual studies of localized changes, and method development without sharing patient scans. They also carry risks. Synthetic scans could be mistaken for real data or used to fabricate evidence, models trained on patient data may memorize training subjects, and a model trained only on an older ADNI cohort may misrepresent other populations. AnaDiffusion is a research model that has not been validated for clinical use, and generated volumes should be labelled as synthetic.

\end{document}